\pdfoutput=1

\documentclass[11pt]{article}

\usepackage[]{EMNLP2022}

\usepackage{times}
\usepackage{latexsym}

\usepackage{arydshln}
\usepackage{subfigure}
\usepackage{amssymb}
\usepackage{multirow}
\usepackage{amsmath} 
\usepackage{color}
\usepackage{mdwlist}

\usepackage{algorithm}
\usepackage{algorithmic}

\usepackage{multirow}
\usepackage{amsmath} 
\usepackage{amssymb}
\usepackage{pifont}
\usepackage{color, soul} 
\usepackage[switch]{lineno}  %
\usepackage{arydshln}
\usepackage{graphicx}
\usepackage{hyperref}

\usepackage[utf8]{inputenc}
\usepackage{cleveref}
\crefname{section}{§}{§}
\crefname{appendix}{Appendix}{§}

\usepackage{hyperref}
\makeatletter
\def\UrlAlphabet{%
      \do\a\do\b\do\c\do\d\do\e\do\f\do\g\do\h\do\i\do\j%
      \do\k\do\l\do\m\do\n\do\o\do\p\do\q\do\r\do\s\do\t%
      \do\u\do\v\do\w\do\x\do\y\do\z\do\A\do\B\do\C\do\D%
      \do\E\do\F\do\G\do\H\do\I\do\J\do\K\do\L\do\M\do\N%
      \do\O\do\P\do\Q\do\R\do\S\do\T\do\U\do\V\do\W\do\X%
      \do\Y\do\Z}
\def\UrlDigits{\do\1\do\2\do\3\do\4\do\5\do\6\do\7\do\8\do\9\do\0}
\g@addto@macro{\UrlBreaks}{\UrlOrds}
\g@addto@macro{\UrlBreaks}{\UrlAlphabet}
\g@addto@macro{\UrlBreaks}{\UrlDigits}

\usepackage{tablefootnote}
\usepackage{longtable}
\usepackage{booktabs}

\usepackage{CJK}

\usepackage[T1]{fontenc}

\usepackage[utf8]{inputenc}

\usepackage{microtype}

\title{MUSIED: A Benchmark for Event Detection from Multi-Source Heterogeneous Informal Texts}

\author{Xiangyu Xi$^{1}$\footnotemark[1] , Jianwei Lv$^{1}$\footnotemark[1] , Shuaipeng Liu$^{1}$\footnotemark[1] \hspace{0.05cm}\footnotemark[2] , Wei Ye$^{2}$\footnotemark[2] , Fan Yang$^{1}$, Guanglu Wan$^{1}$ \\
  $^1$ Meituan Group, Beijing, China \\
  $^2$ National Engineering Research Center for Software Engineering, Peking University, \\Beijing, China \\
   \texttt{liushuaipeng@meituan.com,wye@pku.edu.cn} \\}

\begin{document}
\begin{CJK}{UTF8}{gkai}
\maketitle
\renewcommand{\thefootnote}{\fnsymbol{footnote}}
\footnotetext[1]{The first three authors contributed equally.}
\footnotetext[2]{Corresponding authors.}
\begin{abstract}

Event detection (ED) identifies and classifies event triggers from unstructured texts, serving as a fundamental task for information extraction. Despite the remarkable progress achieved in the past several years,  most research efforts focus on detecting events from formal texts (e.g., news articles, Wikipedia documents, financial announcements). Moreover, the texts in each dataset are either from a single source or multiple yet relatively homogeneous sources. With massive amounts of user-generated text accumulating on the Web and inside enterprises, identifying meaningful events in these informal texts, usually from multiple heterogeneous sources, has become a problem of significant practical value. As a pioneering exploration that expands event detection to the scenarios involving informal and heterogeneous texts, we propose a new large-scale Chinese event detection dataset based on user reviews, text conversations, and phone conversations in a leading e-commerce platform for food service. We carefully investigate the proposed dataset's textual informality and multi-source heterogeneity characteristics by inspecting data samples quantitatively and qualitatively. Extensive experiments with state-of-the-art event detection methods verify the unique challenges posed by these characteristics, indicating that multi-source informal event detection remains an open problem and requires further efforts. 
Our benchmark and code are released at \url{https://github.com/myeclipse/MUSIED}.

\end{abstract}

\maketitle

\section{Introduction}
\label{sec:introduction}


Event detection (ED), which aims to identify event triggers and classify them into specific types from unstructured texts, has been widely researched and applied in various downstream tasks \cite{basile2014extending,cheng-erk-2018-implicit,kuhnle2021reinforcement}. 
Advanced models have been continuously proposed, ranging from feature-based models \cite{liao:2010using,hong-etal-2011-using,li2013joint} to recent neural-based models \cite{chen2015event,nguyen2016joint,chen-etal-2018-collective,xi2021improving,xiangyu-etal-2021-capturing}.
Despite the significant progress, we find that previous works have the following two limitations in practical scenarios.

\textbf{1. Current efforts mainly focused on event detection from formal texts. }  For example, a popular line of works \cite{li2013joint,chen2015event,nguyen2016joint,chen-etal-2018-collective,DBLP:conf/acl/LouLDZC20} aim to detect general domain events from news articles (e.g., ACE 2005 \cite{doddington2004automatic}) and Wikipedia documents (e.g., MAVEN \cite{wang2020maven}). Some other explorations involve extracting events from the financial announcements  \cite{yang2018dcfee,zheng2019doc2edag,liang2021f} or cybersecurity articles\cite{trong2020introducing}, which are also written in a relatively official style. In practical scenarios, however, we usually face the bottleneck of identifying events from informal texts. Compared with formal text, texts produced in more casual contexts (e.g., online chat and phone conversation) pose some unique challenges of long event triggers, high event density, and typos noises, as revealed in our 
analysis (\cref{sec:analysis of informal texts}). Indeed, with vast amounts of user-generated text accumulating on the open Web and private enterprise systems, extracting meaningful events in these informal texts has become an urgent problem of significant practical value.  

\textbf{2. The targeting event-related texts are either from a single source or multiple yet homogeneous sources.}  Most recent datasets (e.g, MAVEN \cite{wang2020maven}, CySecED \cite{trong2020introducing}, ChFinAnn \cite{yang2018dcfee}, and BRAD \cite{lai2021event}) are built from an individual data source. 
The most widely-used ACE 2005 \cite{doddington2004automatic} covers six sources, which are, however, relatively homogeneous internet media to some extent. Regarding informal text, end-users can produce them in many different ways, and hence they have more versatile expressing styles. Therefore, multi-source heterogeneity comes as another difficulty that inherently accompanies text informality. For example, texts generated via online chat and phone calls in after-sales services may greatly diversify, e.g., on length and style. Unfortunately, current ED works fail to adequately address the issue of multi-source heterogeneity.

To address these two problems, in this paper, we expand event detection to the scenarios involving informal and heterogeneous texts. We construct a new large-scale Chinese event detection dataset based on Meituan\footnote{\url{https://about.meituan.com/en}}, 
the most popular Chinese e-commerce platforms for food service, which provides users with multiple ways to feed back on food safety issues (events), such as posting reviews and communicating with after-sale staff. These reviews and conversations yield a large-scale multi-source heterogeneous informal text repository, which contains valuable information about food safety events and hence can serve as a corpus. We collect the desensitized data from three typical scenarios: i) users posting reviews, ii) users communicating with after-sale staff through text messages, and iii) users communicating with after-sale staff on the phone. By extracting user reviews, text conversations, and phone conversations, we create a massive dataset consisting of \underline{MU}lti-\underline{S}ource heterogeneous \underline{I}nformal texts for \underline{E}vent \underline{D}etection (MUSIED).


We investigate MUSIED's textual informality (\cref{sec:analysis of informal texts}) and multi-source heterogeneity  (\cref{sec:analysis of multiple sources}) by carefully inspecting data samples. The textual informality leads to event descriptions involving long triggers (\cref{sec:long_triggers}), multi-event sentences (\cref{sec:multiple_event}), and user typos (\cref{sec:typos}), while the multi-source heterogeneity brings notable diversity of event type distribution and event density across domains (\cref{sec:analysis of multiple sources}).
We re-implement the state-of-the-art ED methods and conduct extensive evaluation on MUSIED (\cref{sec:experiments}). 
The experimental results clearly verify the unique challenges posed by the above characteristics. Specifically, the proposed dataset requires more robust models towards identifying long triggers (\cref{sec:Challenge of Long Triggers}), capturing multi-event interaction (\cref{Challenge of Multi-Event Sentences}), and alleviating typo noises(\cref{error analysis}).
Meanwhile, MUSIED also facilitates future research on tackling multi-source heterogeneity, e.g., with multi-domain learning and (\cref{multi-domain learning}) and domain adaptation (\cref{domain adaptation}).

Our contributions can be summarized as follows:
\begin{itemize}
    \item We expand event detection to the scenarios involving informal and heterogeneous texts, for the first time, by carefully curating a new large-scale dataset.
    \item Extensive experiments with state-of-the-art methods verify the unique challenges posed by textual informality and multi-source heterogeneity characteristics, and indicate multiple promising directions worth pursuing.
\end{itemize}

\section{Event Detection Definition}

We follow the classical settings and terminologies adopted by ACE 2005 program \cite{doddington2004automatic} and MAVEN \cite{wang2020maven}, and specify the vital event terminologies as follows.
\textbf{Event}: a specific occurrence involving participants (location, time, subject, object, etc.). 
\textbf{Event Mention}: a phrase or sentence within which an event is described.
\textbf{Event Trigger}: the main word or phrase that most clearly expresses the occurrence of an event.
\textbf{Event Type}: the semantic class of an event.

Event detection aims to identify event trigger words and classify their event types for a given text. Accordingly, ED is conventionally divided into two subtasks:
(1) \textbf{Trigger identification}, which aims to identify the event triggers.
(2) \textbf{Trigger classification}, which aims to classify the recognized trigger into predefined categories. 
Both subtasks are evaluated with micro precision, recall, and F-1 scores.
Most recent works \cite{chen2015event,nguyen2016joint,chen-etal-2018-collective,wang-etal-2019-adversarial-training} perform trigger classification directly (add an additional type ``N/A'' to be classified at the same time, indicating that the candidate is not a trigger). 
We also inherit these settings in this paper.

\section{Data Collection and Annotation}


\subsection{Data Collection}
\begin{figure}[!hbt]
    \centering
    \includegraphics[scale=0.4]{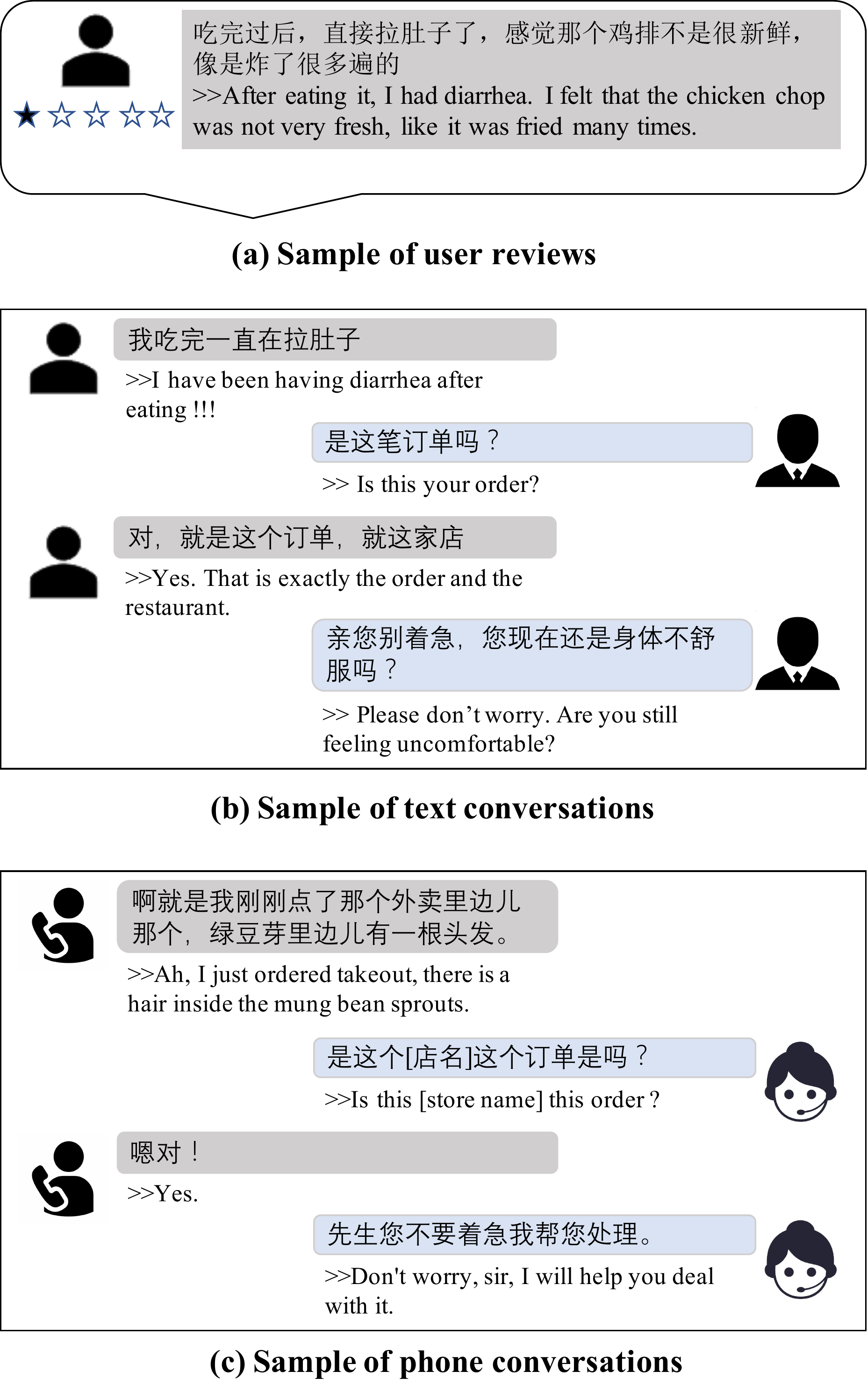}
    \caption{Samples of our corpus.}
    \label{fig:scenarios}
\end{figure}

We collect data from Meituan, which provides users with multiple channels to feed back on food safety issues (events), among which the three most common ways are:
i) users post reviews to restaurants where they have ordered food;
ii) users communicate with after-sale staff through text messages;
iii) users communicate with after-sale staff on the phone.
First, we collect the user reviews, text conversations, and phone conversations from logs of online services for a week.
Further, we desensitized and anonymized the private information from the raw data (see \cref{Ethical Impact} for details).
The samples from each scenario are shown in Figure \ref{fig:scenarios} to promote understanding. 
Note that the phone conversations are speech data, which is transformed into text data via the Automatic Speech Recognition (ASR) service \cite{wang2019overview,kaur2021automatic}.

The above collected data may not involve food safety events (e.g., users make positive reviews). 
We hire annotators to select the reviews and conversations involving food safety incidents. Finally, we retained 4,226 user reviews, 3,767 text conversations, and 3,388 phone conversations, forming a corpus composed of 11,381 documents in total.

\subsection{Event Schema Construction}
\begin{figure}[hbt]
    \centering
    \includegraphics[scale=0.35]{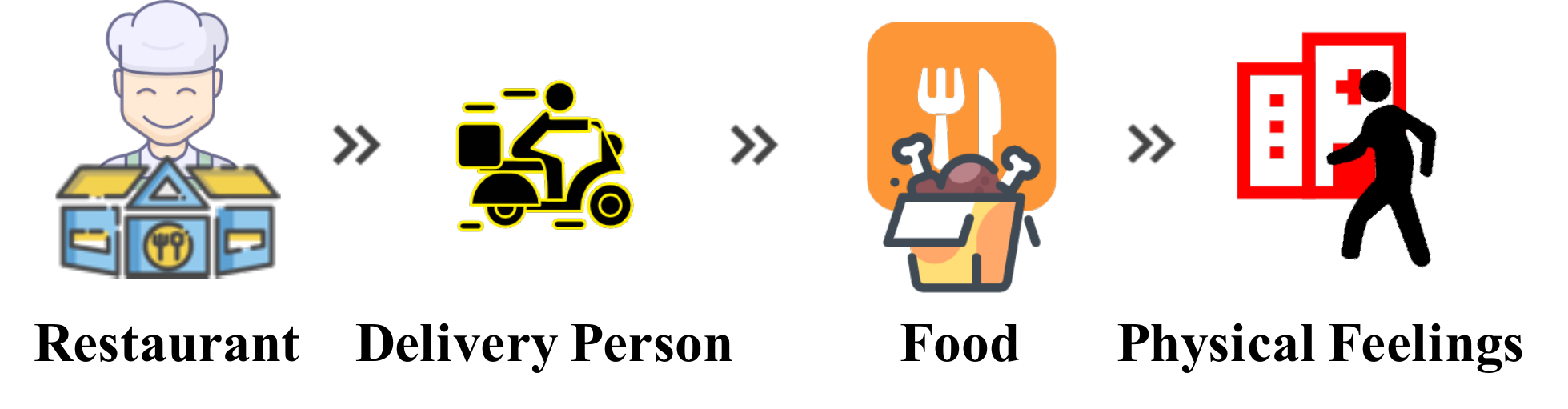}
    \caption{A typical food delivery service scenario.}
    \label{fig:event_schema}
\end{figure}

With the assistance of food safety experts, 
we construct an event schema, from the perspective of users. 
We exemplify using a typical food delivery service scenario shown in Figure \ref{fig:event_schema}, where users usually feed back in terms of:
(1)
\textbf{Food quality} 
    Poor food quality is the main cause of food safety problems (e.g., food is expired or undercooked). 
(2)
\textbf{Restaurant}
    The illegal or improper behaviors of restaurants (e.g., uses illegal food additives) may lead to food safety problems.
(3)
\textbf{Delivery person}
    A small but noticeable percentage of food safety problems are caused by the delivery person (e.g., damages the packaging and pollutes the food).
(4)
\textbf{Physical feelings}
    Rather than above causes, the users may directly express their physical feelings (e.g., feel uncomfortable), which suggest the existence of food safety problems.
Finally, the schema contains 21 event types and broadly covers the user's feedback about above cases. Please refer to \cref{event type schema} for the full event schema description.

\subsection{Data Annotation}

\subsubsection{Annotation Process}
Though with a detailed annotation guideline, the annotation process is complicated and error-prone. 
For accuracy and consistency, we organize a two-stage iterative annotation, following ACE 2005 \cite{doddington2004automatic} and MAVEN \cite{wang2020maven}.
We recruit 20 annotators with food safety domain knowledge, and train them with the guideline. After that, they are given an annotation exercise and 9 annotators with accuracy $>$ 90\% are selected to perform formal annotation.
At the first stage, each document is annotated by 3 independent annotators. The annotation is finished if and only if 3 annotators reach an agreement. Otherwise, in the second stage, all 9 annotators and language experts will discuss documents with annotation disagreements together and determine the final results.

\subsubsection{Annotation Challenges And Solutions} 
\label{annotation_challenge}

\textbf{Candidate Selection}
Since Chinese lacks natural delimiters, words are necessarily generated by segmentation toolkits, which might not exactly match with triggers \cite{DBLP:conf/nlpcc/ZengYFWZ16,lin-etal-2018-nugget}.
Also, the informal texts are more diverse.
It would be impractical and inaccurate to select words with specific features, as done in English dataset \cite{wang2020maven}.
To address above challenge, we annotate in a character-wise manner, instead of performing word segmentation and word-wise annotation sequentially. In this way, though the trigger candidate set is larger because each possible phrase is regarded as a candidate trigger, we tackle the problem of i) limitation of word boundary and 2) error propagation of word segmentation toolkits. 
\\
\textbf{Boundary confusion}
During annotation, we find the triggers are usually followed or surrounded by stop words (such as auxiliary words, modal particles, etc), especially in telephone conversations. We follow the principle that event triggers should not contain redundant information, as long as they can fully express the event information. For example, we do not annotate the modal particles in the following sentence S1. 
``\emph{臭}(\emph{stinky})'' and ``\emph{吃吐}(\emph{Eat and vomit})'' are the triggers of \textit{Abnormalities} and \textit{Uncomfortable} event. However, the token ``\emph{的}'' and ``\emph{了}'' following them are modal particles in Chinese, and do not express useful information.

\textbf{S1}: 
\emph{The duck intestines were \textcolor{red}{stinky}, I \textcolor{blue}{Eat and vomit}.}
(\emph{鸭肠是\textcolor{red}{臭}的，把人都\textcolor{blue}{吃吐}了})
\\
\textbf{Ambiguous User Expression}
The informal user statements are not rigorous and may be insufficient for resolving ambiguities for event types.
For example, for the trigger ``\emph{梆硬}(\emph{hard})'' in the following sentence S2, some annotators believe the reason for ``\emph{梆硬}(\emph{hard})'' is that the chicken is undercooked and considers it as a trigger of \textit{Undercooked} event,
while others think the reason is that the temperature is too low and treats it as a trigger of \textit{Cold} event. 

\textbf{S2}: 
 \emph{I felt that the chicken chop was} \textit{\textcolor{red}{cold}, and the chicken in the chicken roll was also \textcolor{blue}{hard}}
(\emph{感觉鸡排\textcolor{red}{冷}了，鸡肉卷里的鸡肉也是\textcolor{blue}{梆硬}的。})


The annotators are required to disambiguate by integrating contextual information. 
For example, considering the context that the user first complains that the chicken chop is cold (i.e., ``\emph{冷} (\emph{cold})''), the annotators tend to believe 
the following phrase ``\emph{梆硬} (\emph{hard}'') also triggers a \textit{Cold} event.


\subsubsection{Annotation Quality}
With the strict annotation process, our dataset is of high quality. 
For data with annotation disagreement in the first stage, all annotators discuss together and reach agreements (by voting sometimes).
Also, we randomly sample 500 documents without annotation disagreement in first stage, and invite different first-stage annotators to annotate these documents. We measure the inter-annotator agreements of annotation between two annotators with Cohen's Kappa score. The results for trigger and type annotation are 0.83 and 0.82 respectively, which belongs to the Near-perfect agreement range of [0.81, 0.99]. 
The annotated samples are shown in \cref{appendix:annnotated data}.

\section{Data Analysis}

\subsection{Data Size}

Following \citet{wang2020maven}, we show the main statistics of MUSIED and compare with the following datasets in Table \ref{tab:1}:
(1) \textbf{ACE 2005} \cite{walker2006ace}, which is the most wide-used dataset and covers general domain events. 
(2) \textbf{Rich ERE} \cite{mitamura2015overview}, which is provided by TAC KBP competition and contains a series of datasets;
(3) \textbf{MAVEN} \cite{wang2020maven}, which is the largest general domain dataset constructed from Wikipedia and FrameNet; 
(4) \textbf{RAMS} \cite{ebner-etal-2020-multi}, which follows the AIDA ontology and uses Reddit articles. 
(5) \textbf{BRAD} \cite{lai2021event}, which covers Black Rebellions events in African Diaspora; 
(6) \textbf{CySecED} \cite{trong2020introducing}, which is the largest cybersecurity event dataset.
We can observe that our MUSIED is large-scale compared with existing datasets. 
In terms of average instance number per event type, MUSIED
has significant advantage over other datasets (e.g., 1,756 of MUSIED v.s. 707 of MAVEN v.s. 162 of ACE 2005). 
Thus, MUSIED can stably train and benchmark sophisticated neural-based models.

\begin{table*}[hbt]
	\centering
	
	\begin{tabular}{llrrrrrr}
	    \hline
	    \multicolumn{2}{c}{\textbf{Dataset}} & \textbf{\#Doc} & \textbf{\#Tokens} & \textbf{\#Sentences} & \textbf{\#Event Types}& \textbf{\#Events} & \textbf{\#Event Mentions}\\
	    \hline
	    
	    \multicolumn{2}{l}{ACE 2005 English}  & 599 & 303k & 15,789 & 33 & 4,090 & 5,349 \\
	    \multicolumn{2}{l}{ACE 2005 Chinese} & 633 & 321k & 7,269 &33&  2,521 & 3,333 \\
	    \multicolumn{2}{l}{ACE 2005 Arabic} & 403 & 150k & 2,710 &33&  2,267 & 2,270
	    \\
	    
	    \multicolumn{2}{l}{Rich ERE} & 1,272 & 854k & 41,708 & 38 & 29,293 & 38,853 \\
	    \multicolumn{2}{l}{MAVEN}& 4,480 & 1,276k & 49,873& 168 & 111,611 &118,732\\
	    \multicolumn{2}{l}{RAMS}&3,993 &1,218k &44,236 & 139 & 9,124 &9,124\\
	    \multicolumn{2}{l}{BRAD}&151 &172k &5,638  & 12&- &4,259\\
	    
	    \multicolumn{2}{l}{CySecED}&300 &- &290,234  & 30&- &8,014\\

	    \hline
	    \multicolumn{2}{l}{MUSIED}& 11,381 & 7,105k & 315,473 & 21 & 30,940 & 35,313\\
	    
	    \hline

	\end{tabular}
	\caption{Dataset statistics. 
	}
	\label{tab:1}
\end{table*}


\begin{figure}[hbt]
    \centering
    \includegraphics[scale=0.3]{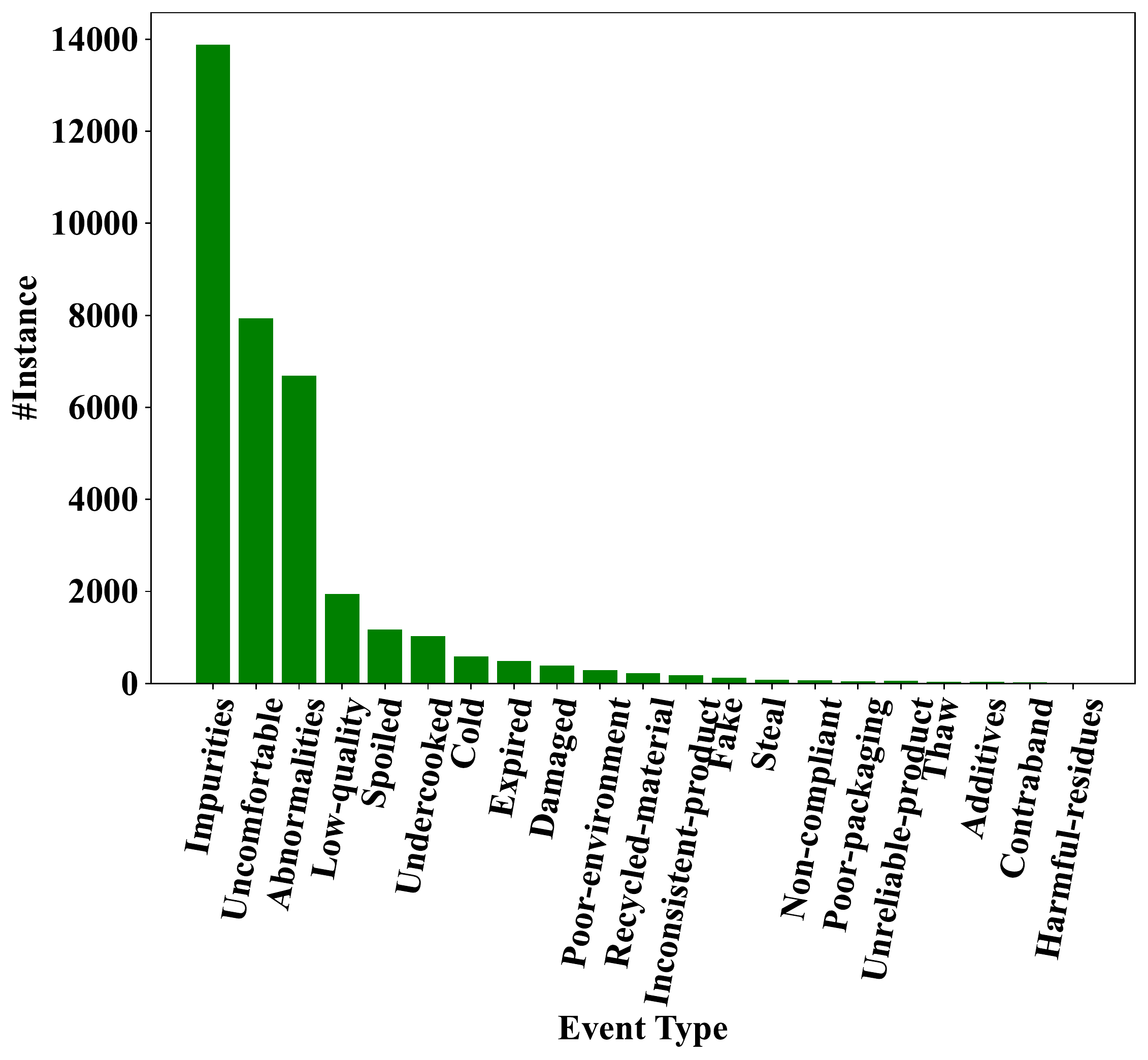}
    \caption{Instance number of each event type.}
    \label{fig:2}
\end{figure}


\subsection{Data Distribution}
\label{sec:data_distribution}

The instance number of each event type is shown in Figure \ref{fig:2}, which shows the existence of the inherent data imbalance problem.
We also display the top 5 event types with their instance numbers and proportions in \cref{top 5 event types}.
MUSIED has 21 event types and 35,313 labeled instances, yet ``Impurities'' (with 13,883 labeled instances) and ``Uncomfortable'' (with 7,935 labeled instances) account for 61.7\% of the data. 
18 (85.7\%) event types have a below-average number of labeled instances and 6 event types even have fewer than 50 labeled instances.
Though potentially hindering the performance of ED models, the occurrence frequency of event types conforms to the long-tail phenomenon in the real world. 
We maintain the original distribution of MUSIED, which can evaluate the ability of the ED models in the long-tail scenario.





\subsection{Analysis of Textual Informality}
\label{sec:analysis of informal texts}
A key characteristic of MUSIED is that the corpus is composed of informal text. 
We introduce the features brought by textual informality as follows.

\subsubsection{Long Triggers}
\label{sec:long_triggers}
Our observation shows that users tend to use more casual expressions and longer triggers to express events. For example, in the following sentence S3, the user says his/her two teeth are broken due to the hard noodles. The phrase ``\emph{牙齿都干掉两颗} (\emph{two teeth are broken})'' triggers an \textit{Uncomfortable} event and consists of 7 tokens.

\textbf{S3:} 
\emph{The rice is rotten, noodles are as hard as steel wire, \textcolor{red}{two teeth are broken}}
(\emph{米饭稀烂，面条跟钢丝条一样硬，\textcolor{red}{牙齿都干掉两颗}})

\begin{figure}[hbt]
    \centering
    \includegraphics[scale=0.3]{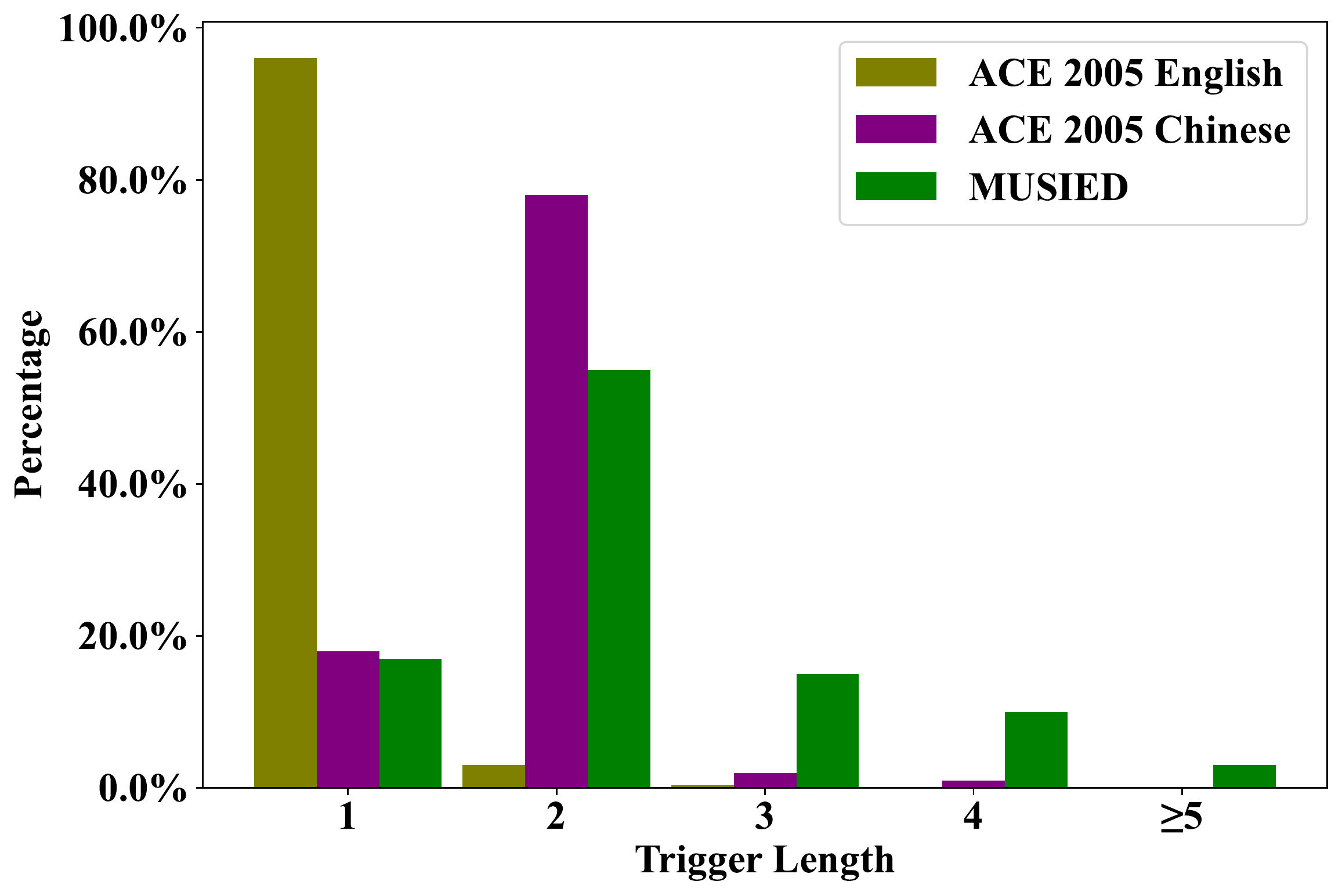}
    \caption{Distribution of triggers with different length.
    }
    \label{fig:tri_len}
\end{figure}

MUSIED contains a much higher proportion of long triggers, as Figure \ref{fig:tri_len} shows. Considering the proportion of triggers consisting of more than 2 tokens, MUSIED is nearly 53 times larger than ACE 2005 English (i.e., 26.97\% v.s. 0.50\%) and 9 times larger than ACE 2005 Chinese (i.e.,  26.97\% v.s. 3.06\%). The long trigger phenomenon poses a great challenge to existing ED models.

\subsubsection{Multiple Events}
\label{sec:multiple_event}
Unlike professionals who write articles or documents in a relatively official style, users may hurriedly express multiple events within one sentence.
For example, in the following sentence S4, the user reports multiple food quality related events, which lead to an \textit{Uncomfortable} event.

\textbf{S4}: 
\emph{Then I ate his fried rice. Because his prawns were \textcolor{red}{not fresh} and \textcolor{blue}{undercooked}. Then I had his grilled sausages and sausages, and it all \textcolor{purple}{didn't feel very fresh}. After eating, I \textcolor{olive}{had diarrhea}.}
(\emph{然后我吃了他那个炒饭因为他那个虾\textcolor{red}{不新鲜}然后也\textcolor{blue}{不熟}然后再加上他那个烤肠啊腊肠啊都是感觉\textcolor{purple}{不是很新鲜}然后吃了之后我我\textcolor{olive}{拉肚子}})


Following previous works \cite{chen2015event}, we make statistics on sentences with multiple events and find that the proportion of multi-events sentence in MUSIED is much larger than ACE 2005 (i.e., 36.9\% of MUSIED v.s. 27.3\% of ACE 2005 English v.s. 19.3\% of ACE 2005 Chinese). The reason lies in that food safety event correlations are closer and users tend to simultaneously express the cause and consequence. 

	

\subsubsection{Typos} 
\label{sec:typos}

Different from formal texts which are produced by professionals, the user-generated informal texts are less rigorous and may contain typos. The automatic speech recognition service may also produce errors.
For example, in the following sentence S5, the user finds the beef rice is sour and expresses a \textit{spoiled} event. 
However, the user types a typo token ``\emph{\textcolor{red}{搜}}'' (means search), which has the same pronunciation (pronounced as ``sou'' in Chinese) but different meaning as the token ``\emph{\textcolor{blue}{馊}}'' (means sour).

\textbf{S5}: 
\emph{I ordered beef rice, it looks \textcolor{red}{search}(\textcolor{blue}{sour}) already}
(\emph{我点的牛肉饭，看起来都\textcolor{red}{搜}(\textcolor{blue}{馊})掉了})

We make statistics on the typos using the state-of-the-art spelling error corrector (SEC) \cite{li2021dcspell}. The result shows that 2.2\% of sentences contain spelling errors, 0.1\% of tokens are typos and 1.5\% of them are within the triggers.
Though the accuracy of SEC may be limited in our corpus, the result together with our observation reveals that existence of typos is a noticeable problem.




\subsection{Analysis of Multi-Source Heterogeneity}
\label{sec:analysis of multiple sources}
In this section, 
we analyze the multi-source heterogeneity from the following perspectives.

\subsubsection{General Textual Features}

The textual features shift remarkably across sources of MUSIED.
We present the statistics on each source in \cref{appendix:domain statistics}, from which we can observe that document size varies significantly (i.e., 1.4 of user reviews v.s. 48.1 of text conversations v.s. 37.8 of phone conversations in terms of \#sentences per document). The reason lies in that conversation with staff is more official and users tend to provide more complete information.
Also, we calculate the average sentence length for each source and further compute the standard deviation of the average sentence lengths.
The standard deviation of MUSIED is notably larger than ACE 2005 (i.e., 5.06 of MUSIED v.s. 3.31 of ACE 2005 English v.s. 3.87 of ACE 2005 Chinese).

\subsubsection{Event Type Distribution and Event Density}

The event type distribution and event density vary significantly across sources of MUSIED. 
The top 5 event types for each source are shown in \cref{top 5 event types}, 
from which we can easily observe the notable diversity of event type distributions across sources of MUSIED.
For a quantitative analysis, we calculate the event type distribution for each source and calculate the wasserstein distance \cite{vallender1974calculation} between the distributions (please refer to \cref{appendix:Event Distribution Statistics} for the detailed calculation procedure).
MUSIED is much larger than ACE 2005 (i.e., $6.17\times 10^{-4}$ of MUSIED v.s. $3.32\times 10^{-4}$ of ACE 2005 English v.s. $1.51\times 10^{-4}$ of ACE 2005 Chinese), in terms of the average wasserstein distance.
Also, we compute the average event density for each source and the standard deviation of the average event densities, which shows that 
the disparity of event density across MUSIED's sources is more remarkable 
(i.e., 0.35 of MUSIED v.s. 0.17 of ACE 2005 English v.s. 0.08 of ACE 2005 Chinese).

To sum up, MUSIED is of more significant heterogeneity and can effectively support the exploration of ED involving multi-source heterogeneity. Conversely, the limited heterogeneity, together with the data scarcity problem, makes ACE 2005 insufficient for benchmarking relevant research.

\section{Experiments}
\label{sec:experiments}

\subsection{Benchmark Settings}
We randomly split the annotated documents into train, dev, and test sets with the ratio of 8:1:1.  The statistics of the three sets are shown in Table \ref{tab:5}.


\begin{table}[hbt]
	\centering
	\begin{tabular}{lrrr}
	    \hline
	    \textbf{Set} &\textbf{\#Doc}&\textbf{\#Sentences} & \textbf{\#Event Mentions}\\
	    \hline
        Train & 9,103 & 252,786 & 28,012 \\
        Dev & 1,139 & 31,269 & 3,540\\
        Test & 1,139 & 31,418 & 3,761\\
        
	    \hline
	\end{tabular}
	\caption{
	The statistics of splitting MUSIED.
	}
	\label{tab:5}
\end{table}


        

\subsection{Experimental Settings}


Recently, neural-based models have achieved significant progress.
Thus, we investigate the following state-of-the-art neural-based methods, which can be roughly divided into two categories:


\textbf{Sentence-Level Models} which use information within the sentence to extract triggers.
\textbf{DMCNN} \cite{chen2015event} which uses CNN as feature extractor and concatenates sentence and lexical feature;
\textbf{BiLSTM} \cite{DBLP:journals/neco/HochreiterS97} which uses bi-directional long short-term memory network as encoder;
\textbf{BiLSTM-CRF} \cite{DBLP:conf/icml/LaffertyMP01} which uses bi-directional long short-term memory network followed by a conditional random field layer;
\textbf{C-BiLSTM} \cite{DBLP:conf/nlpcc/ZengYFWZ16} which proposes a convolution bidirectional LSTM to capture both sentence-level and lexical information;
\textbf{DMBERT} \cite{wang-etal-2019-adversarial-training} which takes BERT as encoder and adopts the dynamic multi-pooling mechanism;
\textbf{BERT} \cite{yang2019exploring} which fine-tune BERT on the down-stream ED task via a sequence labeling manner.

\textbf{Document-Level Models} which integrate the document-level contextual information.
\textbf{HBTNGMA} \cite{chen-etal-2018-collective} which 
dynamically fuses the sentence- and document-level information;
\textbf{MLBiNet} \cite{DBLP:conf/acl/LouLDZC20} which captures the document-level association of events.

The implementation details such as hyperparameters are listed in \cref{appendix:hyperparameters}.
Following previous works, we report \emph{Precision} (P), \emph{Recall} (R) and \emph{F1-Score} (F1) on trigger classification.

\begin{table}[hbt]
	\centering
	
	\begin{tabular}{cccc}
	    \hline
        \textbf{Model}&\textbf{P} & \textbf{R} & \textbf{F1} \\
        \hline
        DMCNN &\textbf{84.2}&56.8 &67.8\\
        BiLSTM &75.6 &66.4 & 70.7 \\
        BiLSTM+CRF &  76.0& 69.8 & 72.8\\
        C-BiLSTM & 75.7 & 70.5 & 73.0\\
        DMBERT  &77.0 &68.7 &72.7\\
        BERT & 72.6 & 78.9& 75.6\\
	    \hline
	    HBTNGMA & 73.1& \textbf{79.5}& \textbf{76.2}\\
	    MLBiNet & 73.4& 69.3 &71.3 \\
	    \hline
	\end{tabular}
	\caption{Performance on trigger classification (\%).}
	\label{tab:4}
\end{table}




\subsection{Overall Experimental Results}
\label{Overall Experiments}

The overall experimental results are shown in Table \ref{tab:4}, from which we have the following observations:
(1) Sequence labeling methods have advantages over token-level classification models. For example, BiLSTM and BERT achieve 2.9 and 2.9 F1 improvements over DMCNN and DMBERT respectively. The reason lies in that token-level classification models separately predict trigger candidates without considering the event interdependency, while sequence labeling methods generate representation and make predictions collectively. 
(2) BiLSTM+CRF achieves notable improvements over BiLSTM (e.g., 72.8 v.s. 70.7 in terms of F1), with the assistance of CRF layer modeling event correlations. The observation confirms our analysis in \cref{sec:multiple_event} that modeling event correlations is important for MUSIED, due to the multi-event sentences.
(3) By incorporating document-level contextual information, HBTNGMA gains an absolute improvement of 3.4 F1-Score over BiLSTM+CRF (i.e., 76.2 v.s. 72.8).
The experiment result is consistent with our observation of ambiguous user expression (\cref{annotation_challenge}), and clearly indicates the importance of document-level contextual information.

\subsection{Analysis of Textual Informality}

\subsubsection{Challenge of Long Triggers}
\label{sec:Challenge of Long Triggers}
As \cref{sec:long_triggers} shows, MUSIED contains long triggers, due to the informal expressions.
We make statistics on BERT's recall on triggers of different lengths, as Table \ref{tab:recall} shows, from which we can easily observe a general trend that
the longer the trigger, the worse the recall rate. 
Existing ED models have difficulty in capturing the distribution pattern of long triggers, and the challenge should be further addressed.

\begin{table}[hbt]
    \centering
    \begin{tabular}{lrrr}
    \hline
    \textbf{Length} & [1,2] & [3,4] &  [5,)\\
    \hline
    \textbf{Recall} & 80.5 & 79.6 & 34.0 \\
    \hline
    \end{tabular}
    \caption{BERT's recall of triggers of different lengths.}
    \label{tab:recall}
\end{table}


\subsubsection{Challenge of Multi-Event Sentences}
\label{Challenge of Multi-Event Sentences}

Following \citet{chen2015event}, we divide the test set into two parts according to the event number in a sentence (single event (i.e., 1/1) and multiple events (i.e., 1/N)), and perform evaluation separately.
From Table \ref{tab:multi-events} we can observe that:
(1) All models perform much worse on 1/N, which coincides with previous findings \cite{chen2015event,chen-etal-2018-collective}.
(2) Though achieving comparable performance in 1/1 data, sequence labeling methods have significant advantage over token-level classification methods on 1/N data (i.e., 42.6 of DMCNN v.s. 60.4 of BiLSTM, 55.1 of DMBERT v.s. 72.1 of BERT). The experimental results indicate that it is worth exploring more collectively-detecting methods, to better capture the distribution pattern of multiple events within a sentence.

\begin{table}[hbt]
	\centering
	\begin{tabular}{lccc}
	    \hline
	    \textbf{Model}&\textbf{1/1} & \textbf{1/N}  & \textbf{All}\\
	    \hline
	    DMCNN &  79.1 & 42.6 & 67.8\\
        BiLSTM &  79.6 & 60.4& 70.7\\
        DMBERT &  82.4 &  55.1 & 72.7\\
        BERT &  84.3& 72.1 & 75.6\\
	    \hline
	\end{tabular}
	\caption{F1-Scores on Single Event Sentences (\textbf{1/1}) and Multiple Event Sentences (\textbf{1/N}).}
	\label{tab:multi-events}
\end{table}

\subsubsection{Challenge of Typos}
\label{error analysis}
We use the state-of-the-art spelling error corrector (SEC) \cite{li2021dcspell} on the test set, then manually collect the samples that are indeed typos.
Further, we retest these corrected samples with BERT, as Table \ref{tab:comparison of typos} shows. 
After correction, some mislabeled samples can be fixed and the performance is improved. 
For example, the S5 in \cref{sec:typos} can be correctly predicted. Another concrete case is shown in \cref{appendix:Case Study of Spelling Error Corrector} to promote understanding.

\begin{table}[hbt]
    \centering
    
    \begin{tabular}{lccc}
    \hline
    \textbf{Sampled Instances} & \textbf{P} & \textbf{R} & \textbf{F1}\\
    \hline
    -w/o Correction & 63.6 & 41.2 & 50.0 \\
    -w/ Correction & 66.7 & 47.1 & 55.2 \\
    \hline
    \end{tabular}
    \caption{Performance on corrected samples.}
    \label{tab:comparison of typos}
\end{table}

\begin{table*}[bth]
	\centering
	
	\begin{tabular}{lcccccccccccc}
	    \hline
	    \multirow{2}{*}{\textbf{Model}} & \multicolumn{3}{c}{\textbf{User Reviews}}&\multicolumn{3}{c}{\textbf{Text Conversations}}&\multicolumn{3}{c}{\textbf{Phone Conversations}}&\multicolumn{3}{c}{\textbf{ALL}} \\
	    \cline{2-13} 
	    & \textbf{P} & \textbf{R} & \textbf{F1} & \textbf{P} & \textbf{R} & \textbf{F1} & \textbf{P} & \textbf{R} & \textbf{F1} & \textbf{P} & \textbf{R} & \textbf{F1}\\
	    \hline
	    SD & \textbf{70.4} & 72.9 & 71.6 & \textbf{81.1} & 79.4 & 80.3 & 65.9 & 66.5 & 66.2 & 72.9 &72.9 &72.9 \\
	    PD & 67.6& 74.7 & 71.0&78.0&85.3 &81.0  &68.4 &73.7 &71.0 & 72.6 & 78.9& 75.6\\
	    PDMT  &67.5 &76.6 &71.8 &76.5 &85.1 &80.6 & 69.0& \textbf{77.2}& \textbf{72.9}&72.1 &80.5 &76.1 \\
	    MDSP &\textbf{70.4} &\textbf{77.9} &\textbf{74.0} & 77.3& \textbf{87.1}& 81.9& \textbf{70.3}& 74.2 &72.2 &\textbf{73.6} & \textbf{80.6} & \textbf{76.9}  \\
	    \hline
	\end{tabular}
	\caption{Performance of BERT with different multi-domain learning strategies (\%).}
	\label{tab:mdl}
\end{table*}

Though of great potential to address the typo challenge, our sampling statistics show the precision of SEC is quite limited in our corpus (47.8\%). One possible reason is the textual features of our corpus are quite different from the SEC's training corpus.
We believe that developing a SEC more suitable for MUSIED and exploring more sophisticated methods such as incorporating pronunciation features may be useful to address the challenge.

\subsection{Analysis of Multi-Source Heterogeneity}
Since the different sources of MUSIED have diversified data characteristics, we investigate the multi-source heterogeneity via the following two typical research topics (i.e., multi-domain learning and domain adaptation). Following \citet{pradhan-etal-2013-towards,ganin2015unsupervised,chen-etal-2021-data,wang-etal-2020-multi-domain-named}, we treat each source as a single ``domain'' in the following investigation.


\subsubsection{Analysis of Multi-Domain Learning}
\label{multi-domain learning}


So far, we exploit a standard strategy by naively pooling all available data across domains (sources) and discarding the domain information.
A shared model is trained to serve all domains. 
However, the multi-source heterogeneity drives us to explore ways to utilize the domain information.
Following \citet{chen-cardie-2018-multinomial,wang-etal-2020-multi-domain-named}, we select BERT and experiment with the following multi-domain learning strategies:
(1) \textbf{SingleDomain (SD)} which trains an individual ED model for each domain separately and only uses the training data for the single domain. 
(2) \textbf{PoolDomain (PD)} which is the strategy we used. The model ignores domain information, albeit uses all available training data.
(3) \textbf{PoolDomain-MultiTask (PDMT)} which is similar to PoolDomain, except that we add an auxiliary task that learns domain labels. The domain information is utilized, though in a simple way.
(4) \textbf{MultiDomain-Shared-Private (MDSP)} which uses 
i) a shared MLP for all domains that extracts generic and domain-invariant features; and
ii) a private MLP for each domain that extracts domain-specific characteristics. 


We report the performance in each domain and overall test set in Table \ref{tab:mdl}, from which we can observe that:
(1) The difficulty of event detection varies across domains. Text conversations is the easiest, and phone conversations is the hardest.
(2) PD outperforms SD, which is consistent with the observations in \citet{chen-cardie-2018-multinomial}. The information sharing between domains may improve the generalization ability of ED models.
(3) PDMT gains slight improvement over PD by utilizing the domain information via a simple multi-task way, demonstrating that domain information can bring effective clues.
(4) Further, the MDSP strategy generally outperforms all models (e.g., achieving 76.9 F1). The shared-private framework can effectively capture common language features shared across domains, as well as domain-specific patterns.
The above analysis show that domain information is effective enhancement, and multi-domain learning deserves more research efforts.

\subsubsection{Analysis of Domain Adaptation}
\label{domain adaptation}

Domain adaptation is another key criteria for evaluating ED models.
Following \citet{naik2020towards}, we investigate the typical unsupervised domain adaptation (UDA) problem, and adopt the following strategies:
(1) \textbf{BERT-Naive} which utilizes the labeled source domain dataset and ignores the target domain data.
(2) \textbf{BERT-ADA} which incorporates the adversarial domain adaptation (ADA) framework to construct representations predictive for ED, but not predictive of the domain. 




\begin{table*}[hbt]
    \centering
    \begin{tabular}{llcccccc}
    \hline
    \multirow{2}{*}{\textbf{Setting}} & \multirow{2}{*}{\textbf{Model}} &
    \multicolumn{3}{c}{\textbf{In-Domain}} & \multicolumn{3}{c}{\textbf{Out-Of-Domain}} \\
    \cline{3-8}
    & & \textbf{P} & \textbf{R} & \textbf{F1} & \textbf{P} & \textbf{R} & \textbf{F1} \\
    \hline
    \multirow{2}{*}{U$\rightarrow$T} & BERT-Naive & 70.4 & 72.9 & 71.6& 65.1& 64.0 & 64.6\\
    & BERT-ADA & 74.6& 74.9& 74.7& 63.0& 64.6& 63.8\\
    \hline
        \multirow{2}{*}{U$\rightarrow$P} & BERT-Naive & 70.4 & 72.9 & 71.6& 59.8 & 60.9 & 60.3\\
    & BERT-ADA & 77.4& 74.7& 76.0& 62.1& 62.2& 62.2\\
    \hline
        \multirow{2}{*}{T$\rightarrow$U} & BERT-Naive & 81.1 & 79.4 & 80.3& 68.7 & 60.0 & 64.1\\
    & BERT-ADA & 79.1& 80.8& 79.9& 70.4& 58.8& 64.1\\
    \hline
        \multirow{2}{*}{T$\rightarrow$P} & BERT-Naive & 81.1 & 79.4 & 80.3& 70.4 &61.7 & 65.7\\
    & BERT-ADA & 81.9& 51.9& 63.5& 70.2& 46.4& 55.9\\
    \hline
        \multirow{2}{*}{P$\rightarrow$U} & BERT-Naive & 65.9 & 66.5 & 66.2& 43.7 &44.6 &44.1\\
    & BERT-ADA & 60.8& 62.7& 61.7& 52.9& 47.3& 49.9\\
    \hline
        \multirow{2}{*}{P$\rightarrow$T} & BERT-Naive & 65.9 & 66.5 & 66.2& 65.7 &65.5 &65.6\\
    & BERT-ADA & 60.2& 62.9& 61.5& 64.5& 65.0& 64.8\\
    \hline

    \end{tabular}
    \caption{Performance of unsupervised domain adaptation on trigger classification (\%). A$\rightarrow$B denotes that A and B are source and target domain. U, T and P denotes user review, text conversations and phone conversations respectively. The performances on both source (i.e., the In-Domain column) and target domain test set (i.e., the Out-Of-Domain column) are reported.}
    \label{tab:uda}
\end{table*}

As Table \ref{tab:uda} shows, we select source and target domain from the three domains in turn, forming six UDA settings. 
Though the ADA framework is reported of advantage \cite{naik2020towards}, it is not the case with MUSIED. BERT-ADA underperforms BERT-Naive in several settings (e.g., U$\rightarrow$T, T$\rightarrow$P and P$\rightarrow$T), which indicates that domain adaptation in MUSIED is challenging due to the multi-source heterogeneity, and more research efforts are required. 
Other DA settings (e.g., semi-supervised DA) can also be effectively supported by MUSIED and should be further investigated.

\section{Related Work}


Most existing works towards event detection adopt general domain datasets such as ACE 2005 \cite{walker2006ace}, TAC KBP datasets \cite{mitamura2015overview} and MAVEN \cite{wang2020maven} as benchmarks.
Also, some works present domain-specific datasets and valuable explorations.
For example, event extraction from biomedical texts are extensively researched \cite{pyysalo2007bioinfer,thompson2009construction,buyko2010genereg,nedellec-etal-2013-overview}.
\citet{sims-etal-2019-literary} present a new dataset of literary events.
CASIE \cite{satyapanich2020casie} and CySecED \cite{trong2020introducing} are proposed to facilitate the research of detecting cybersecurity events.
Continuous works \cite{yang2018dcfee,zheng2019doc2edag,liang2021f} have focused on detecting financial events from the Chinese financial announcements (i.e., ChFinAnn dataset).
\citet{lai2021event} presents BRAD, focusing on Black Rebellions events in African Diaspora.



However, most existing works focus on detecting events from formal texts (e.g., news articles, Wikipedia documents, etc), and target the datasets where the texts are either from a single source (e.g., MAVEN \cite{wang2020maven}, CySecED \cite{trong2020introducing}, ChFinAnn \cite{yang2018dcfee}) or multiple yet homogeneous sources (e.g., ACE 2005 \cite{doddington2004automatic}).
In this paper, we present a massive multi-source heterogeneous informal text dataset for event detection, for the first time.
It is also the first food safety event detection dataset.

\section{Conclusion and Future Work}

We have presented MUSIED, a massive multi-source heterogeneous informal text dataset for event detection, based on user reviews, text conversations and phone conversations of online food services.
The extensive evaluation verify the unique challenges posed by the textual informality and multi-source heterogeneity characteristics.
Our in-depth investigations present multiple promising directions worth pursuing, including exploiting document-level information, multi-domain learning and domain adaptation.
In the future, we are interested in extending MUSIED to more event-related tasks such as event argument extraction.

\section*{Acknowledgement}
We thank the annotators for their efforts into data annotation, and for their continuous verification and revision after the submission of the paper.
We also thank Yang An for his remarkable assistance with the spelling error corrector experiments (\cref{sec:typos} and \cref{error analysis}) and Yuncheng Hua for helpful discussions.
This research was supported by the National Key Research and Development Program of China (No. 2021YFC3340101).

\clearpage

\section*{Limitations}
MUSIED is composed of Chinese corpus, which might be less friendly to researchers who are unfamiliar with Chinese.
However, considering many non-English datasets have been proposed and promoted research in related fields (e.g., Douban Conversation Corpus \cite{wu-etal-2017-sequential} in dialogue system, DuReader \cite{he-etal-2018-dureader} in machine reading comprehension, etc.), we believe that the language barrier does not hinder the contribution of MUSIED to the community. Also, we provide a well-documented homepage and easy-to-use toolkits including preprocessing, models and checkpoints, to further reduce the impact of language barrier.

\section*{Ethics Impact}
\label{Ethical Impact}
In consideration of ethical concerns, we provide the following detailed description:
\begin{enumerate}
    \item The corpus is sampled from the logs of a real e-commerce platform, and we strictly desensitized and anonymized the private information. Following \citet{chen2020jddc}, we mask the sensitive information including user's phone number, user's name, user's address, restaurant's name, restaurant's address, etc (e.g. replacing phone number with special token <PHONE-NUMBER>, and replacing order IDs with <ORDER-ID>). The dataset \textbf{does not contain} any personally identifiable information, sensitive personal data, or commercially sensitive data.
    \item The dataset has been collected in a manner which is consistent with the terms of use. The data officer of the e-commerce platform has authorized us to collect and open source the dataset. The dataset is freely accessible online without copyright constraint to academic use.
    \item We hired 20 annotators with food safety domain knowledge and paid them with a fair salary (i.e.,35 dollars per hour) during the annotation. The annotators are treated fairly and able to give informed consent.
\end{enumerate}

\section*{Broader Impact}
For the first time, we expand event detection to the scenarios involving informal and heterogeneous texts, by carefully curating a new large-scale dataset.
In this paper, our extensive experiments with state-of-the-art methods verify the unique challenges posed by textual informality and multi-source heterogeneity characteristics, and indicate multiple promising directions worth pursuing.
We believe our work can inspire broader investigation in the future.



\bibliography{anthology}
\bibliographystyle{acl_natbib}

\appendix
\clearpage




    

\section{Event Type Schema}
\label{event type schema}
We present the event type schema along with their descriptions in Table \ref{tab:event_description}.
The schema contains 21 event types and broadly covers the user's feedback about food quality, restaurant, delivery person, and user's physical feelings.

\section{Data Analysis}
\subsection{Top 5 Event Types}
\label{top 5 event types}
We show the top 5 event types along with their instance number and proportion in Table \ref{tab:top_5}. The top 5 event types for both overall corpus (i.e., the {\tt ALL} row) and each domain (i.e., the {\tt User Review}, {\tt Text Conversation} and {\tt Phone Conversation} rows) are shown.

\begin{table}[hbt]
	\centering
	\begin{tabular}{lrr}
	    \hline
	    \textbf{Event Type} & \textbf{\#Event Mentions} & \textbf{Proportion}\\
	    \hline
	    \multicolumn{3}{c}{{\tt ALL}} \\
	    \hdashline
        \textit{Impurities} & 13,883 & 39.3\% \\
        \textit{Uncomfortable} & 7,935 & 22.5\% \\
        \textit{Abnormalities} & 6,684 & 18.9\%\\
        \textit{Low-quality} & 1,950 & 5.5\%\\
        \textit{Spoiled} & 1,176 & 3.3\%\\
        \hline
        \multicolumn{3}{c}{{\tt User Review}} \\
        \hdashline
        \textit{Abnormalities} & 1,865& 38.1\% \\
        \textit{Impurities} & 783& 16.0\% \\
        \textit{Undercooked} & 651 &  13.3\%\\
        \textit{Uncomfortable}& 541 &  11.0\%\\
        \textit{Cold} & 414 &  8.4\%\\
        \hline
        \multicolumn{3}{c}{{\tt Text Conversation}} \\
        \hdashline
        \textit{Impurities} & 6,696 & 42.2\% \\
        \textit{Uncomfortable} & 3,579 & 22.6\% \\
        \textit{Abnormalities} & 2,510 &  15.8\%\\
        \textit{Low-quality} & 1,757 &  11.1\%\\
        \textit{Spoiled} & 476 &  3.0\%\\
        \hline
        \multicolumn{3}{c}{{\tt Phone Conversation}} \\
        \hdashline
        \textit{Impurities} & 6,404 & 43.9\% \\
        \textit{Uncomfortable} & 3,815 & 26.2\% \\
        \textit{Abnormalities} & 2,309 &  15.9\%\\
        \textit{Spoiled} & 649 &  4.4\%\\
        \textit{Expired} & 272 &  1.9\%\\
        \hline

	    \hline
	\end{tabular}
	\caption{Statistics of top 5 event types.}
	\label{tab:top_5}
\end{table}

\subsection{Domain Statistics}
\label{appendix:domain statistics}
For each domain of MUSIED, we present the detailed statistics in Table \ref{tab:domain_statistics}. 

\begin{table*}[hbt]
	\centering
	\begin{tabular}{lrrrrr}
	    \hline
	     \textbf{Domain}&\textbf{\#Document} & \textbf{\#Tokens} & \textbf{\#Sentences} & \textbf{\#Events} & \textbf{\#Event Mentions}\\
	    \hline
	    User Review & 4,226 & 144k & 6,083 & 4,036 & 4,898\\
         Text Conversation & 3,767 & 3,136k & 181,316 & 14,686 & 15,858\\
        
        Phone Conversation & 3,388 & 3,805k & 128,074 & 12,218 & 14,557\\
	    \hline
	    Total & 11,381 & 7,105k & 315,473 & 30,940 & 35,313\\
	    \hline
	\end{tabular}
	\caption{Domain statistics of MUSIED. }
	\label{tab:domain_statistics}
\end{table*}

\subsection{Statistics of Event Type Distribution}
\label{appendix:Event Distribution Statistics}
Given a corpus with $N$ domains and $M$ event types, we first calculate the event type distribution $P_i$ for each domain $i$ as follows:
\begin{equation}
\begin{aligned}
    P_i &= (p_{i,1},p_{i,2},...,p_{i,M}) \\
    p_{i,j} &= \frac{\#(\mbox{triggers with type $j$ in domain $i$})}{\#(\mbox{triggers with any type in domain $i$})}
\end{aligned}
\end{equation}
where $p_{i,j}$ denotes the occurrence frequency of type $j$ in domain $i$.
$\#(\mbox{triggers with type $j$ in domain $i$})$ denotes the number of triggers with type $j$ in domain $i$. $\#(\mbox{triggers with any type in domain $i$})$ denotes the number of triggers in domain $i$.

Then, we calculate the wasserstein distance between the event type distributions of any two domains (exemplify with $P_1,P_2$) as follows:
\begin{equation}
    W(P_1,P_2) = inf_{\gamma \sim \Pi( P_1, P_2)} E_{(x,y) \sim y} [|x-y|]
\end{equation}

Further, we calculate the average wasserstein distance as follows:
\begin{equation}
\begin{aligned}
    \bar W = \frac{1}{M}\frac{1}{C_N^2}\sum_{i=1}^{N}{\sum_{j=i+1}^{N}{W(P_i,P_j)}} 
\end{aligned}
\end{equation}
where $C_N^2 = \frac{N(N-1)}{2}$ denotes the number of domain pairs and $M$ denotes the number of event types.

\subsection{Word-Trigger Mismatch}


As a Chinese dataset, MUSIED lacks natural delimiters and also suffers from the word-trigger mismatch problem existing in ACE 2005 Chinese dataset \cite{DBLP:conf/nlpcc/ZengYFWZ16,lin-etal-2018-nugget,xiangyu2019hybrid}. The words generated by word segmentation toolkits might not exactly match with event triggers. Following \citet{xiangyu2019hybrid}, we make statistics on two types of word-trigger mismatch: i) Cross-word Triggers where a trigger might be composed of multiple words; ii) Inside-word Triggers where a single character or some consequent characters inside a word can be a trigger.
The statistical results with three different word segmentation tools are shown in Table \ref{tab:mismatch}, from which we can observe that proportion of problematic triggers in MUSIED is much larger than ACE 2005 Chinese dataset (i.e., 35.24\% v.s. 16.15\%).
The severe word-trigger mismatch problem poses a great challenge and may hinders the performance of word-wise event detection models.

\begin{table*}[ht]
	\centering
	\begin{tabular}{lcccccc}
	    \hline
	    \multirow{2}{*}{\textbf{Toolkits}}&\multicolumn{3}{c}{\textbf{ACE 2005 Chinese}} & \multicolumn{3}{c}{\textbf{MUSIED}}\\
        \cline{2-7}
        &\textbf{C-W} & \textbf{I-W} & \textbf{R} & \textbf{C-W} & \textbf{I-W} & \textbf{R}\\
        \hline
        CoreNLP\tablefootnote{https://nlp.stanford.edu/software/segmenter.shtml} & 2.25\%&11.79\% &85.96\% & 26.21\% & 11.38\%&  62.41\%\\
        Jieba\tablefootnote{https://github.com/fxsjy/jieba} &2.31\% & 17.94\% &79.75\% & 23.06\%& 9.35\% & 67.59\%\\
        NLPIR\tablefootnote{https://github.com/NLPIR-team/NLPIR} & 8.97\% & 5.19\% & 85.84\%& 29.58\% & 5.98\% & 64.44\%\\
        Average&4.51\% &11.64\% &83.85\% & 26.28\%& 8.96\% & 64.76\%\\
	    \hline
	\end{tabular}
	\caption{Statistics of word-trigger mismatch. C-W, I-W and R denotes cross-word triggers, inside-word triggers and regular triggers respectively.}
	\label{tab:mismatch}
\end{table*}

\section{Samples of Annotated Data}
\label{appendix:annnotated data}
To promote understanding, we show the sample of annotated data from three domains, as Table \ref{tab:annotated_data} shows.

\begin{table}[hbt]
    \centering
    \begin{tabular}{p{0.9\columnwidth}}
    \hline
    Sample \# 1 Domain: User Review \\
    \hdashline
    The dishes have \textcolor{red}{hair}, the restaurant does not reply to us, I often order dishes in this restaurant. \\
    菜里/有/\textcolor{red}{毛发}，跟/商家/沟通/也/不/回复，我/还/经常/点/他们/家/外卖。\\
    \hdashline
    Event Trigger: \textcolor{red}{毛发}(\textcolor{red}{hair}) \\
    Event Type:  Impurities \\
    \hline
    \hline
    Sample \# 2 Domain: Text Conversation \\
    \hdashline
    Last time I shopped, the noodles \textcolor{red}{expired} \\
    上次/买/东西/面/就/\textcolor{red}{过期}/了 \\
    \hdashline
    Event Trigger: \textcolor{red}{过期}(\textcolor{red}{expired}) \\
    Event Type: Expired \\
    \hline
    \hline
    Sample \# 3 Domain: Phone Conversation \\
    \hdashline
    So why does that fried chicken have \textcolor{red}{black splots}? \\
    所以/那份/炸鸡/为什么/会有/\textcolor{red}{黑斑}？\\
    \hdashline
    Event Trigger: \textcolor{red}{黑斑}(\textcolor{red}{black splots}) \\
    Event Type: Spoiled \\
    \hline
    \end{tabular}
    \caption{Sample of annotated data from three domains.}
    \label{tab:annotated_data}
\end{table}

\section{Hyperparameters}
\label{appendix:hyperparameters}
In this section, we introduce the hyperparameter settings and training details of various ED models that we implemented for experiments.

\subsection{BERT-based Models}
For both \textbf{DMBERT} and \textbf{BERT}, we use the BERT$_{\rm BASE}$ for Chinese, and the released pre-trained checkpoints can be downloaded at \url{https://storage.googleapis.com/bert_models/2018_11_03/chinese_L-12_H-768_A-12.zip}. Adam with learning rate of 2e-05, $\beta_1=0.9$ and $\beta_2=0.999$ is used for optimization. We set the training epochs and batch size to 50 and 64 respectively. We set dropout to 0.1.

\subsection{LSTM-based Models}
For \textbf{BiLSTM}, \textbf{BiLSTM+CRF} and \textbf{C-BiLSTM}, we use the pre-trained Chinese word embeddings \footnote{\url{https://github.com/Embedding/Chinese-Word-Vectors}}. The adopted hyperparameters are shown in Table \ref{tab:hyper-lstm}.

\begin{table}[hbt]
	\centering
	\begin{tabular}{l|c}
	    \hline
	    Epoches & 50\\
        Batch Size & 64\\
	    Dropout Rate & 0.1\\
	    Learning Rate & 2e-05\\
	    Dimension of Word Embedding & 300\\
	    Dimension of Hidden Layers & 300\\
	    Layers of LSTM & 1\\
	    Kernel Size of CNN & 3\\
	    Number of Feature Map & 300\\
	    Optimizer & Adam\\
	    \hline
	\end{tabular}
	\caption{Hyperparameter settings for the BiLSTM-based models.}
	\label{tab:hyper-lstm}
\end{table}

\subsection{DMCNN model}
For DMCNN, we use the pretrained Chinese word embeddings, and the hyperparameters are shown in Table \ref{tab:hyper-dmcnn}.

\begin{table}[hbt]
	\centering
	\begin{tabular}{l|c}
	    \hline
	    Epoches & 50\\
        Batch Size & 64\\
	    Dropout Rate & 0.1\\
	    Learning Rate & 2e-05\\
	    Dimension of Word Embedding & 300\\
	    Kernel Size of CNN & 3\\
	    Number of Feature Map & 300\\
	    Optimizer & Adam\\
	    \hline
	\end{tabular}
	\caption{Hyperparameter settings for the DMCNN model.}
	\label{tab:hyper-dmcnn}
\end{table}

\subsection{HBTNGMA model}
For HBTNGMA, We use the official code released by \citet{chen-etal-2018-collective} \footnote{\url{https://github.com/yubochen/NBTNGMA4ED}}.
We adopt the original hyperparameters from \citet{chen-etal-2018-collective} except that we use the pretrained Chinese word embeddings with 300 dimension.


\subsection{MLBiNet model}

For MLBiNet, we use the official code released by \citet{DBLP:conf/acl/LouLDZC20} \footnote{\url{https://github.com/zjunlp/DocED}}.
We use the pretrained Chinese word embeddings with 300 dimension.



\section{Computing Issues}
The computing issues are explained in this section.

\textbf{Computing Infrastructure} We implemented our model with TensorFlow v1.4.0 and Pytorch v1.7.0, and trained our models on NVIDIA Tesla v100 GPU. The operation system is CentOS 7.6.

\textbf{Computational Budget} 
Table \ref{tab:runtime} shows the used computing infrastructures and the average running time per epoch of various models.

\begin{table}[hbt]
	\centering
	\begin{tabular}{ccc}
	    \hline
	    \multirow{2}{*}{\textbf{Model}} & \textbf{Computing} &  \multirow{2}{*}{\textbf{Runtime}} \\
	    & \textbf{Infrastructure} & \\
        \hline
        DMCNN& 1$\times$ Tesla v100 &40 min \\
        BiLSTM & 1$\times$ Tesla v100 & 5 min \\
        BiLSTM+CRF & 1$\times$ Tesla v100 & 8 min \\
        C-BiLSTM &1$\times$ Tesla v100 & 10 min\\
        DMBERT &1$\times$ Tesla v100 & 55 min \\
        BERT & 1$\times$ Tesla v100 &30 min\\
	    HBTNGMA & 1$\times$ Tesla v100 &20 min\\
	    MLBiNet & 1$\times$ Tesla v100 &20 min \\
	    \hline
	\end{tabular}
	\caption{The average runtimes per epoch of various models.}
	\label{tab:runtime}
\end{table}

\section{Experimental Results}

\subsection{Overall Performance}
Following previous works \cite{li2013joint,chen2015event}, we only report \emph{Precision} (P), \emph{Recall} (R) and \emph{F1-Score} (F1) on trigger classification task in \cref{Overall Experiments}. The performance on trigger identification task is also shown in Table \ref{tab:identification_and_classification}.

\begin{table}[hbt]
	\centering
	\begin{tabular}{cccc}
	    \hline
        \textbf{Model} &\textbf{P} & \textbf{R} & \textbf{F1} \\
        \hline
        DMCNN&\textbf{85.5} &57.6 &68.8 \\
        BiLSTM & 77.1&67.7 &72.1 \\
        BiLSTM+CRF &77.5 & 70.1 & 73.6 \\
        C-BiLSTM & 77.2 & 71.9 & 74.5\\
        DMBERT &79.4 &70.8 &74.9 \\
        BERT & 73.8 & \textbf{80.3} & 76.9\\
	    \hline
	    HBTNGMA &74.3 & \textbf{80.3} &\textbf{77.2}\\
	    MLBiNet &75.5 & 72.3& 73.9 \\
	    \hline
	\end{tabular}
	\caption{Overall performance on trigger identification.}
	\label{tab:identification_and_classification}
\end{table}

\subsection{Case Study of Spelling Error Corrector}
\label{appendix:Case Study of Spelling Error Corrector}
A concrete case S7 is shown to demonstrate the benefit of Spelling Error Corrector (SEC). The user intends to express that he/she feels unwell (``\emph{感觉/不\textcolor{blue}{适}}(\emph{feel unwell})''). However, the user types a typo token ``\emph{\textcolor{red}{是}}'' (means yes), which has the same pronunciation (pronounced as ``shi'' in Chinese) but different meaning as the token ``\emph{\textcolor{blue}{适}}'' (means physically well). The word ``\emph{不\textcolor{red}{是}}'' is a widely-used statement of expressing negation. All models fail to recognize the instance due to the typo before correction, and can fix the error with correction.

\textbf{S7}: \emph{豆腐面条鸡蛋然后吃了之后身体就感觉不\textcolor{red}{是} (\textcolor{blue}{适})} (\emph{Tofu, noodles and eggs. After eating them, I feel \textcolor{red}{not} (\textcolor{blue}{unwell})})

\subsection{Impact of Data Imbalance}
\label{sec:Challenge of Data Imbalance}
As \cref{sec:data_distribution} shows, the inherent data imbalance problem exists in MUSIED.
To quantitatively investigate the effect, we first rank labels (i.e., event types) based on the number of their corresponding training instances and then divide them into several subsets with continuous rankings. 
Since instances with a specific label may be too few, empirical results on instances of a label set could yield more robust and convincing conclusions.
The first event type alone forms a single subset, and the remaining 20 event types are equally grouped into three subsets. In this way, we finally get a division of four subsets, named \emph{Subset-1}, \emph{Subset-2}, \emph{Subset-3} and \emph{Subset-4}, which contain 1, 6, 7 and 7 labels respectively.

\begin{table}[hbt]
	\centering
	\small
	\begin{tabular}{lcccc}
	    \hline
	    \textbf{Model}&\emph{Subset-1} & \emph{Subset-2}  & \emph{Subset-3}& \emph{Subset-4}\\
	    
	    \hline
	    DMCNN &   82.39\%& 55.67\%& 41.21\%& 13.32\%\\
        BiLSTM &  87.45\% & 61.46\%& 48.24\%& 10.51\%\\
        DMBERT & 82.96\%& 61.42\%& 42.71\%& 23.52\%\\
        BERT &  88.96\%& 64.38\%& 59.31\%& 49.99\%\\
	    \hline
	\end{tabular}
	\caption{F1-score of different models in four subsets.}
	\label{tab:data-imbalance}
\end{table}

As Table \ref{tab:data-imbalance} shows, we collect the F1-scores of four baselines for each subset, from which we can find that the data imbalance problem significant hinders the performance and results in a degradation (e.g., 88.96 of \emph{Subset-1} v.s. 64.38 of \emph{Subset-2} v.s. 59.31 of \emph{Subset-3} v.s. 49.99 of \emph{Subset-4} for BERT). The performance is significantly worse when label has fewer training instances.
Hence, further explorations on handling the data imbalance challenge may be critical for MUSIED.

\begin{table*}[hbt]
	\centering
	\begin{tabular}{llp{1.3\columnwidth}}
	    \hline
	   \textbf{ID} & \textbf{Event Type} & \textbf{Description} \\
	    \hline
	    & \textbf{Restaurant} & The illegal or improper behaviors of restaurants lead to food safety problems. \\ 
	    \hdashline
        1 & \textit{Additives} & 
         Restaurant uses illegal food additives, including food additives with irregular labels and unknown sources. \\
        \hdashline
        2 & \textit{Contraband}&  
        Restaurant sells commodities that are prohibited or contains non-food raw materials \\
        \hdashline
        3 & \textit{Harmful-residues}&  
        Restaurant sells food that contains harmful residues, such as pesticide, biological toxins, and heavy metals. \\
        \hdashline
        4 & \textit{Poor-environment}& 
        Restaurant provides unsanitary dinning environments.\\
        \hdashline
        5 & \textit{Recycled-material}& 
        Restaurant sells food that is produced using recycled food as raw material. \\
        \hdashline
        6 & \textit{Inconsistent-product}& 
         Restaurant sells food that is inconsistent with the advertisement, such as food quantity, dish content, etc.
        \\
        \hdashline
        7 & \textit{Fake}& 
        Restaurant sells fake food with counterfeit, shoddy, or unauthorized materials. \\
        \hdashline
        8 & \textit{Low-quality}& 
        Restaurant is reported by users to have unspecified food quality problems. \\
        \hdashline
        9 & \textit{Non-compliant}& 
         
        Restaurant provides service in non-compliant status, including
        1) without a license, 2) with fake licenses, and 3) the scope of licenses does not match the actual scope of business. \\
        \hdashline
        10 & \textit{Poor-packaging}& 
         Restaurant provides poor food packaging (or dinnerware), such as simple, thin and smelly packaging with non-food-graded materials.
        \\
        \hdashline
        11 & \textit{Unreliable-product}& 
        Restaurant sells food that contains products without a production date, quality certificate, or manufacturer's source. \\
        \hline
        & \textbf{Delivery Person} & The illegal or improper behaviors of delivery person lead to food safety problems. \\
        \hdashline
        12 & \textit{Damaged}& 
        Delivery person damages or pollutes the food packaging, which affects the quality of food or ingredients.  \\
        \hdashline
        13 & \textit{Steal}& 
        Delivery person is suspected to steal (part of) food based on the quantity and packaging integrity.\\
        \hline
        & \textbf{Food Quality} & The poor food quality lead to food safety problems. \\
        \hdashline
        14 &  \textit{Spoiled}& Food or ingredients have obviously deteriorated, moldy, or rotten, both internally and externally. \\
        \hdashline
        15 & \textit{Undercooked}& 
        Food or ingredients are undercooked.\\
        \hdashline
        16 & \textit{Cold} &
        Food or ingredients have low temperatures, which affects the taste. \\
        \hdashline
        17 & \textit{Expired}& 
        Food or ingredients are expired. \\
        \hdashline
        18 & \textit{Thaw}&  
        Food or ingredients are melting due to improper cold chain distribution. \\
        \hdashline
        19 & \textit{Impurities}& 
        Food or ingredients contains undesirable and disgusting objects, such as eggshells, hair, etc.  \\
        \hline
        & \textbf{Physical Feelings} & The users' physical feelings suggest the existence of food safety problems. \\
        \hdashline
        20 & \textit{Uncomfortable}& 
        User feels unwell after the meal, in terms of physical feelings. \\
        \hdashline
        21 & \textit{Abnormalities}& 
        User feels unwell with the meal, in terms of visual or gustatory feelings. \\
        
	    \hline
	\end{tabular}
	\caption{The 21 event types in MUSIED and their corresponding descriptions.}
	\label{tab:event_description}
\end{table*}

\end{CJK}
\end{document}